\documentclass[10pt,twocolumn,letterpaper]{article}

\usepackage{cvpr}              %

\definecolor{cvprblue}{rgb}{0.21,0.49,0.74}
\usepackage[pagebackref,breaklinks,colorlinks,allcolors=cvprblue]{hyperref}
\usepackage{multirow}
\usepackage{multicol}
\usepackage{adjustbox}
\usepackage{soul}
\usepackage{wrapfig}
\usepackage{makecell}
\usepackage{natbib}
\setcitestyle{square, comma, numbers,sort&compress}
\usepackage{booktabs}
\usepackage{array}
\usepackage{colortbl}
\usepackage{xcolor}

\def\paperID{15175} %
\def\confName{CVPR}
\def\confYear{2025}

\title{There is no SAMantics! \\ Exploring SAM as a Backbone for Visual Understanding Tasks}

\author{Miguel Espinosa*\thanks{miguel.espinosa@ed.ac.uk}, Chenhongyi Yang*, Linus Ericsson, Steven McDonagh, Elliot J. Crowley\\
School of Engineering, University of Edinburgh\\
{\tt\small }
}

\begin{document}

\maketitle

\makeatletter
\newcommand{\ourmethod}{
    \@ifnextchar. {RefSAM}
        {\@ifnextchar, {RefSAM}
            {\@ifnextchar: {RefSAM}
              {RefSAM~}
            }
        }
    }
\makeatother

\newcommand{\apbox}{AP$^{bb}$}
\newcommand{\apmask}{AP$^{mk}$}

\begin{figure*}[h]
    \begin{center}
        \includegraphics[width=0.85\textwidth]{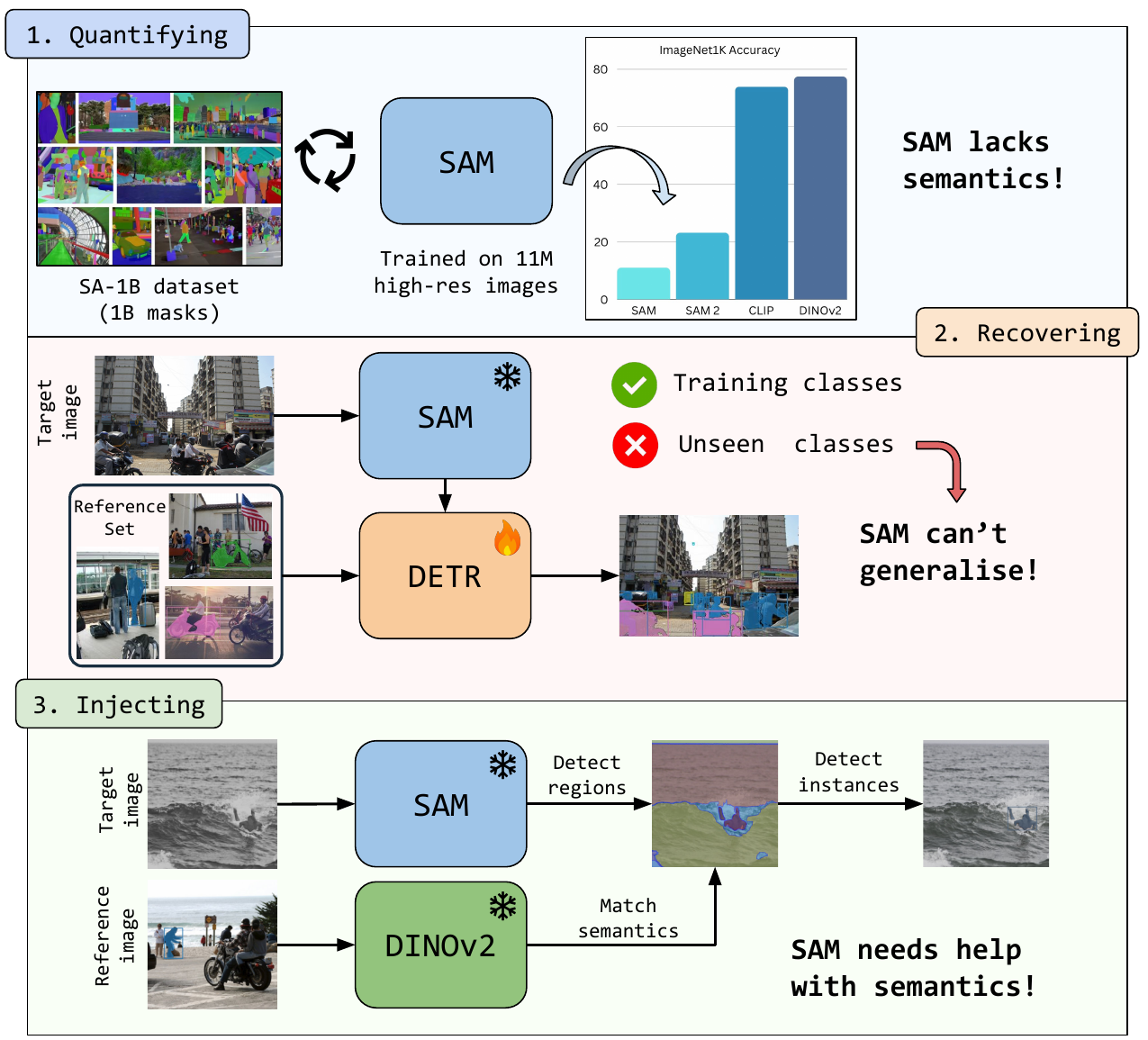}
    \end{center}
    \caption{Exploring SAM's Semantic Gap for Image Understanding. (1) \textbf{Quantifying SAM’s Semantic Understanding}: Despite training on a very large dataset, SAM lacks inherent semantics, as shown by its lower ImageNet1K classification accuracy compared to CLIP and DINOv2 models. (2) \textbf{Recovering Semantics with Fine-tuning}: SAM's ability to generalise remains limited; it can identify classes in the training set but struggles with unseen classes even with in-context learning through DETR. (3) \textbf{Injecting Semantics from External Models}: By integrating semantic-rich representations from models like DINOv2, we can enhance SAM's ability to match semantics and improve its understanding of segmented regions.}
    \label{fig:teaser}
\end{figure*}

\begin{abstract}

The Segment Anything Model (SAM) was originally designed for label-agnostic mask generation. Does this model also possess inherent semantic understanding, of value to broader visual tasks? In this work we follow a multi-staged approach towards exploring this question. %
We firstly quantify SAM's semantic capabilities by comparing base image encoder efficacy under classification tasks, in comparison with established models (CLIP and DINOv2). Our findings reveal a significant lack of semantic discriminability in SAM feature representations, limiting potential for tasks that require class differentiation. This initial result motivates our exploratory study that attempts to enable semantic information via in-context learning with lightweight fine-tuning where we observe that generalisability to unseen classes remains limited. Our observations culminate in the proposal of a training-free approach that leverages DINOv2 features, towards better endowing SAM with semantic understanding %
and achieving instance-level class differentiation through feature-based similarity. Our study suggests that incorporation of external semantic sources provides a promising direction for the enhancement of SAM's utility with respect to complex visual tasks that require semantic understanding.

\texttt{Code:} \url{https://github.com/miquel-espinosa/samantics}

\end{abstract}

\section{Introduction}

Large vision models~\cite{clip, dinov2, deit, vit}, trained on extensive image datasets~\cite{sam}, are powerful tools in computer vision. Among these, the Segment Anything Model (SAM)~\cite{sam} stands out for its impressive image segmentation capabilities. SAM and its extension to video data, SAM 2~\cite{sam2}, allow a user to easily segment an input image into its different regions and objects. This interaction supports multiple prompt types---such as points, bounding boxes, text, and rough masks---to produce accurate label-agnostic pixel-wise masks.
SAM has greatly accelerated many annotation workflows for image-based segmentation tasks, significantly reducing manual burden~\cite{gallagher2024multispectral, chen2023fine, kulkarni2024anytime, jiang2023segment, chen2023segment}.

However, SAM’s functionality comes with a %
fundamental limitation:~\textbf{it operates without any inherent understanding of \textit{what} it is segmenting}. Despite its extensive training on diverse image content, SAM remains unable to classify or semantically interpret the image regions that it segments. While it can reliably delineate objects, SAM does so without differentiating between classes, such as distinguishing a cat from a dog or a car from a tree.

Semantic understanding is key for many segmentation related applications. Many tasks require %
more than just %
the delineation of object boundaries. For instance, automatic labelling of in-the-wild datasets requires not only identifying regions, but also assigning meaningful class labels without manual intervention. Other high-level visual understanding tasks, such as instance segmentation, object detection, and classification, require semantic understanding to determine the class of an instance. Without semantics, models like SAM lack the capacity to assign \emph{interpretable} labels, limiting their utility in real-world applications that depend on a deeper understanding of scene content.

This gap---the absence of semantics---raises questions about the missing components that limit SAM’s interpretive abilities. As an analogy, we observe that large language models trained through self-supervised prediction of missing or masked words allows them to implicitly learn contextual and relational meanings. By exploiting large datasets such models learn how language components fit together naturally, enabling the emergence of a form of semantic understanding. However, we will evidence that SAM’s segmentation prowess does not extend to classification in similar fashion. What underlying challenges prevent it from achieving a semantic understanding of segmented regions? 
Can a model trained only to predict label-agnostic masks on large-scale datasets implicitly learn semantic understanding of the objects it segments? %
How can SAM’s segmentation ability be extended to provide true semantic understanding of segmented objects?

In this work, we explore SAM's semantic limitations and investigate the potential to enhance SAM by injecting semantic understanding into the architecture, towards increasing the utility and value of such large unified segmentation frameworks. To this end, we undertake a multi-stage study:%

\begin{enumerate}
    \item \textbf{Quantifying SAM’s Semantic Understanding.} We benchmark SAM feature representations against popular vision encoders (CLIP~\cite{clip} and DINOv2~\cite{dinov2}) on image classification tasks, measuring the presence of semantics via linear probing on respective features. Despite impressive segmentation abilities, we find SAM lacks discriminative feature quality to enable successful classification, underscoring limited semantic encoding.
    \item \textbf{Recovering Semantics in SAM.} We then explore whether SAM’s inherent representations can be adapted for semantic tasks through lightweight fine-tuning. By introducing in-context learning with reference images, we help SAM capture class-specific information for more effective segmentation. This approach shows some success but reveals a critical limitation: the adapted model overfits to known classes, struggling to generalise to new ones.
    \item \textbf{Injecting External Semantics.} Motivated by SAM’s inherent semantic gap, we experiment with injecting semantics directly from a pretrained, semantically rich model. Using DINOv2 as a backbone, we create a hybrid architecture that fuses SAM’s segmentation strengths with external semantic knowledge. Preliminary results suggest %
    a promising direction for the integration of %
    semantic awareness, without exhaustive model retraining.
\end{enumerate}

Our experimental observations highlight that while SAM can achieve limited semantic insight when tuned for specific object categories, the lack of generalisation ability remains a critical obstacle. 
Through this investigation, we provide insight into enhancing large segmentation models with semantic understanding and present promising routes for bridging the current semantic gap.

\section{Related Work}
\label{sec:related-work}
\textbf{Image Segmentation.}
The idea of automatically generating contours, or segmentation masks, for images has been a popular topic in the computer vision community~\cite{sam,MaskRCNN,unet,fcn,segmenter,mask2former}, and has uses within various fields, including medical imaging (e.g. tumour and organ segmentation)~\cite{unet} and autonomous driving (e.g. identifying pedestrians, vehicles and road signs)~\cite{drivingsegment}. There are different ways of formulating the problem. The task of \textit{semantic segmentation} is to, given an input image, predict the class at each pixel location. \textit{Instance segmentation} differs in that, rather than simply assigning a class to each pixel, each individual instance of a class needs to be distinguished and labeled separately, so that multiple objects of the same class (e.g., two people) are represented as distinct entities.

Convolutional neural networks (CNNs)~\cite{fcn} quickly became the norm for segmentation approaches, with the seminal U-Net~\cite{unet} architecture becoming widely used due to its encoder-decoder structure that preserves spatial details while capturing context. Meanwhile two-stage methods for object detection~\cite{RCNN,FastRCNN,FasterRCNN} were enhanced with instance segmentation capabilities through Mask R-CNN~\cite{MaskRCNN} or the transformer-based DETR~\cite{detr}. More recently, the introduction of Vision Transformers (ViTs)~\cite{vit} has led to transformer-based segmentation models like Segmenter~\cite{segmenter} and Mask2Former~\cite{mask2former}. These models leverage self-attention mechanisms to capture long-range dependencies within images, achieving state-of-the-art results by integrating global context. SAM (Segment Anything Model)~\cite{sam} is an impressive  transformer-based segmentation model trained not to perform traditional semantic segmentation (or instance segmentation) as we have described. It instead used a dataset without class annotations to learn highly effective segmentation capabilities without any of the semantic understanding. It is this lack of semantics that we focus on in this work.

\textbf{Reusing SAM.}
The Segment Anything Model (SAM) has been widely adopted due to its versatility, serving as support for numerous tasks. Several methods have leveraged SAM to extend to tasks like image editing \cite{yu2023inpaint,xie2023edit,yao2024matte,Zhang2023May_personalise_SAM}, object tracking \cite{yang2023track} and weakly supervised segmentation \cite{chen2023segment,ru2023token} by using its masks as prior information. However, SAM faces challenges in complex scenarios, where its category-agnostic nature struggles to differentiate between multiple instances or semantically rich regions.

One way of dealing with this is to retrain a new model from scratch e.g.\ using SA-1B~\cite{sam} for a custom task~\cite{t_rex2, pan2023tap}. Another is to retrain SAM on secondary datasets with rich labels (e.g.\ COCO, ADE20K)~\cite{li2023semantic}. Other work directly reuses the existing SAM model. Training-free approaches~\cite{Zhang2023May_personalise_SAM, Liu2023May_matcher} exploit feature-matching methodologies and correlations with frozen representations. Yet, they end up being complicated pipelines with manually defined thresholds~\cite{Liu2023May_matcher}, or only work under constrained scenarios~\cite{Zhang2023May_personalise_SAM}.~\cite{zhu2024unleashing} pair SAM with Stable Diffusion for open-vocabulary segmentation, and~\cite{lai2023lisa} use language models for text-based instructions. %
Other approaches learn to optimise spatial and semantic prompts on the query space \cite{sun2024vrp} and on the embedding space \cite{Huang2024Jan_learning_to_prompt}. However, they prove insufficient when dealing with multi-instance scenarios.%

Overall, these adaptations show SAM’s potential when combined with semantic signals but highlight its limitations in handling multi-instance, semantically complex tasks, and context-sensitive applications.

\textbf{In-context learning in vision models.}
In-context learning \cite{brown2020language,zhao2021calibrate} enables models to adapt to different tasks with just a few prompts or examples. Specifically, in-context segmentation prompting \cite{zou2023seem,sun2024vrp} aims to utilise a segmented reference image to guide the segmentation process in a target image.
Work such as \cite{Bar2022Sep_visual_icl, Painter, Sun2023Apr_exploring_effective_factors} explores image inpainting at inference time on MAE-pretrained \cite{he2022masked} models, allowing the unification of multiple visual tasks into a single inpainting problem~\cite{SegGPT}. A different approach is explored in~\cite{Bai2023Dec_sequential_modeling} by constructing visual features to minimise cross-entropy loss for next token prediction. T-Rex2 \cite{t_rex2} uses contrastive learning to merge text prompts and visual prompts, and DINOv~\cite{li2023visual_icl} tackles both referring and open-set segmentation with the same framework. SINE~\cite{liu2024simple} specifically tackles task ambiguity for in-context segmentation simply by solving three problems (identical object segmentation, semantic segmentation, instance segmentation) with the same generalist framework. Other methods\cite{zou2023xdecoder,zou2023seem} aim to unify multiple tasks by sharing the same decoder model. Pretrained diffusion models, which encode high quality spatial information~\cite{zhan2023does,zhang2024tale}, have also been explored for image prompting by effectively making use of their text embeddings~\cite{Khani2023Oct_slime, Nguyen2023Nov_visual_instr_inversion}.

In this work, we aim to adapt SAM for in-context semantic prompting as a means to test its underlying semantics.

\section{Preliminaries}

\textbf{SAM Model.} SAM~\cite{sam} consists of three primary components: an image encoder, a prompt encoder, and a mask decoder. The image encoder is a pretrained Vision Transformer (ViT)~\cite{vit} that is adapted for high-resolution inputs~\cite{li2022exploring}. The prompt encoder processes both sparse prompts (e.g.\ points, bounding boxes, text) and dense prompts (e.g.\ rough masks), adding positional encodings~\cite{tancik2020fourier}, to the learnt embeddings. The mask decoder efficiently generates masks by using a modified Transformer decoder block \cite{detr} with self-attention and cross-attention on the prompts. SAM’s training objective incorporates both a focal loss~\cite{focal_loss} and dice loss~\cite{dice_loss} to balance mask accuracy and boundary quality. Importantly, there is no semantic supervision in its training.

\textbf{SAM 2 Model.} SAM 2~\cite{sam2} extends SAM’s segmentation abilities to video. The model employs a streaming approach, using a pre-trained Hiera~\cite{ryali2023hiera} encoder to generate multiscale feature embeddings for each frame. Temporal consistency is maintained through memory attention, where each frame’s features are conditioned on past frames and prompts using cross-attention. A memory bank stores embeddings for recent frames and prompted frames, alongside object pointers for high-level object information. The prompt encoder, similar to SAM, accepts clicks, boxes, or masks, while the mask decoder handles ambiguities by predicting multiple masks across frames and resolving object presence when needed.

\section{Quantifying Semantics in SAM}
\label{sec:quantify-semantics}

To quantify the presence of semantics in SAM we benchmark it against popular visual encoders (CLIP~\cite{clip} and DINOv2~\cite{dinov2}) for image classification through linear probing.

With this, we determine whether SAM, originally designed for label-agnostic mask generation, can be effective for tasks that require semantic differentiation.

\subsection{SAM as an Encoder for Classification Tasks}

We evaluate the feature representations learnt by the SAM and SAM 2 image encoders through a linear probing approach on image classification tasks. The goal is to assess how well the image encoders of these models capture meaningful semantic information when compared to other established encoders, CLIP~\cite{clip} and DINOv2~\cite{dinov2}.

\textbf{Setup.} In both SAM and SAM 2, the image encoder is separated from the mask decoder to isolate the learnt feature representations. SAM employs a Vision Transformer (ViT) backbone for high-resolution image encoding, while SAM~2 uses a hierarchical vision transformer (Hiera) backbone for multi-scale feature processing. For classification, we freeze the parameters of each image encoder and attach a linear classification head. This head is trained on a labelled dataset to classify images based on the encoder's feature representations.

\textbf{Linear Probing.} Linear probing is a common approach to evaluate pretrained representations. Given an image encoder $f$ that maps input images $x$ to feature representations $z=f(x)$, we attach a linear classifier $g$ on top of $f$ to predict class labels. The classifier is trained to minimise a cross-entropy loss 
$L = -\sum y \log(g(z))$ where $y$ is the true class label. By keeping the encoder frozen and only training $g$, we assess how separable the features $z$ are for different classes, reflecting the quality of the representations learned during SAM’s mask-focused training.

{\renewcommand{\arraystretch}{1.2} %
\begin{table}[h]
\centering
\begin{tabular}{lcc}
\toprule
\textbf{Model} & \textbf{Top-1 Acc. (\%)} & \textbf{Top-5 Acc. (\%)} \\ \midrule
SAM            & 11.06 & 25.37 \\
SAM 2          & 23.16 & 44.44 \\
CLIP           & 73.92 & 92.89 \\
DINOv2         & 77.43 & 93.78 \\ \bottomrule
\end{tabular}
\caption{Image classification accuracy of SAM, SAM 2, CLIP, and DINOv2 on the ImageNet1K dataset using linear probing. 
}
\label{tab:imagenet-classification}
\end{table}
    
}

\textbf{Training Details.} We train the linear classifier head on top of the frozen image encoder of each model. The classifier is trained for 10 epochs with a batch size of 128, a learning rate of $10^{-3}$, and mixed precision (fp16). These settings are consistent across all backbone experiments to ensure fair comparison. Input resolutions are standardised to 1024 for SAM models, while CLIP and DINOv2 use a resolution of 224, following their original training configurations.

\textbf{Results on ImageNet1K.} Table~\ref{tab:imagenet-classification} presents the Top-1 and Top-5 classification accuracies achieved by SAM, SAM 2, CLIP, and DINOv2 on the ImageNet1K dataset using linear probing. 
Results reveal a significant performance gap between SAM models and standard image encoders. SAM models show substantially lower accuracy, indicating a lack of inherent semantic understanding typically required for effective image classification.

\textbf{Discussion.} These findings underscore that training on label-agnostic mask generation, even at scale, does not endow SAM models with the semantic discriminability seen in image classification models. Instead, SAM models focus on generalised segmentation without differentiating between specific object classes, which ultimately limits their effectiveness for tasks requiring semantic knowledge.

\section{Recovering Semantic information in SAM via in-context learning}
\label{sec:recovering-semantics}

Having quantified the lack of semantics in the SAM image encoder, we now explore its ability to recover implicit semantics---after seeing vast amounts of images during its pretraining phase---by adapting the model for in-context learning with lightweight finetuning.

\subsection{In-context segmentation prompting}
\label{sec:in-context-segmentation-prompting}
In-context segmentation prompting aims to segment a target image by leveraging a reference segmented image as an example (rather than relying on geometric prompts). More formally, given a reference image $I_r$ and its associated annotations $M^i_r$ for each object category $i$, the goal is to utilise this information to segment the corresponding regions in a target image $I_t$ that belongs to the same category $i$. Here, reference images $I_r$ are also referred to as in-context examples. 
This framework allows for a more flexible and context-sensitive approach to segmentation, as it enables the model to generalise beyond purely geometric inputs.

\subsection{Adapting SAM}

\begin{figure*}[h]
\begin{center}
\includegraphics[width=\textwidth]{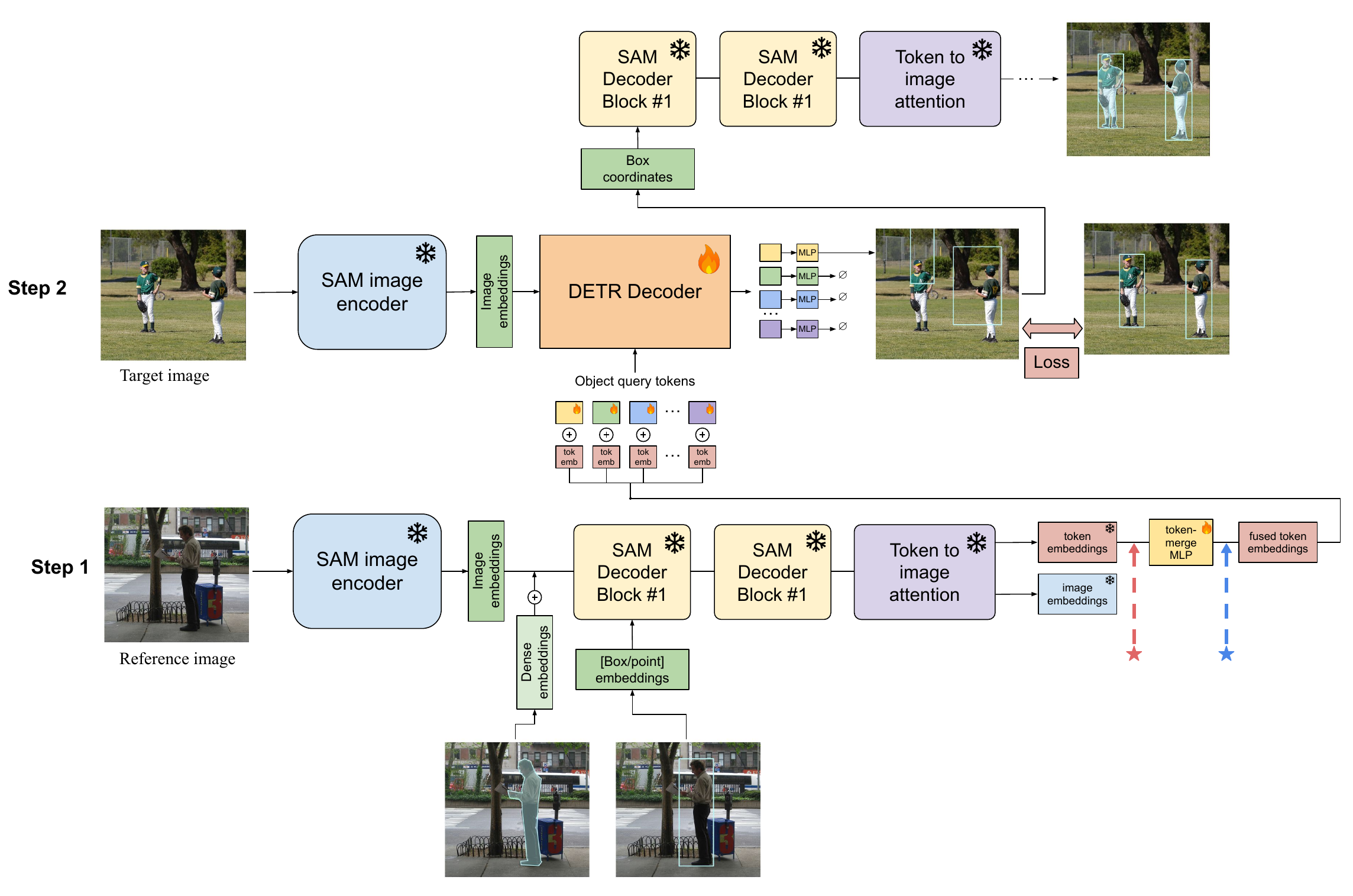}
\end{center}
   \caption{End-to-end training pipeline for in-context segmentation prompting.  First, a reference image is encoded with SAM, together with its corresponding category annotation to obtain token embeddings. Second, we encode the target image and condition the DETR decoder head on the reference token embeddings to generate box proposals. To obtain the final masks we use the predicted boxes to condition SAM to generate mask predictions. SAM encoder and decoder are completely frozen and reused, only the DETR head and the lightweight token-merge MLP layers are trained.}
\label{fig:sam-incole}
\end{figure*}

To adapt SAM for in-context learning, we propose an architecture that largely reuses the base knowledge already present in the pretrained model. Specifically, we freeze both encoder and decoder and solely train a DETR~\cite{detr} decoder head to bridge the semantics of the reference image with the target images. This way, we hope to minimise visual feature discrepancy from reference and target since both representations come from the same frozen model. By using a DETR decoder we natively accept multi-instance images.

As shown in Figure~\ref{fig:sam-incole}, the output embeddings from the SAM decoder for the reference image are added to the DETR query tokens. Thus, the DETR head learns to predict object instances in the target image given the reference image query tokens. Finally, the predicted bounding boxes are used as prompt for the SAM decoder, together with the target image SAM encodings to generate instance masks.

{\begin{table}[h]
\centering
\setlength{\tabcolsep}{1pt} %
\renewcommand{\arraystretch}{1.2} %

\begin{tabular}{lcccccc}
\toprule
\multicolumn{7}{c}{\textbf{COCO}} \\
\midrule
& \textbf{AP$_{\text{box}}$} & AP$^{50}_{\text{box}}$ & AP$^{75}_{\text{box}}$ & \textbf{AP$_{\text{mask}}$} & AP$^{50}_{\text{mask}}$ & AP$^{75}_{\text{mask}}$ \\
\midrule
Adapted SAM (b) & 37.3 & 54.9 & 39.8 & 31.8 & 51.2 & 33.3 \\
Adapted SAM (l) & 43.5 & 61.7 & 47.1 & 37.5 & 58.5 & 40.2 \\
\bottomrule
\end{tabular}
\caption{Performance on COCO validation split on box AP and mask AP. We run on two different model sizes b: base and l: large.}
\label{tab:coco}
\end{table}
}
{\begin{table}[h]
\centering
\setlength{\tabcolsep}{1pt} %
\renewcommand{\arraystretch}{1.2} %

\begin{tabular}{llcccccc}
\toprule
\multicolumn{8}{c}{\textbf{COCO BASE vs NOVEL}} \\
\midrule
& & \textbf{AP$_{\text{box}}$} & AP$^{50}_{\text{box}}$ & AP$^{75}_{\text{box}}$ & \textbf{AP$_{\text{mask}}$} & AP$^{50}_{\text{mask}}$ & AP$^{75}_{\text{mask}}$ \\
\midrule
\multirow[b]{2}{*}{\textbf{\makecell{Adapted \\ SAM}}} & \makecell{BASE \\ classes}  & 24.2 & 30.1 & 25.6 & 21.2 & 29.0 & 23.9 \\
& \makecell{NOVEL \\ classes} & 7.1 & 9.7 & 7.2 & 6.9 & 10.1 & 7.9 \\
\bottomrule
\end{tabular}
\caption{Performance on COCO BASE vs NOVEL validation splits (see Figure~\ref{fig:coco-novel-images} for a full list of the category splits). Results highlight the performance gap between BASE and NOVEL classes.}
\label{tab:coco-novel}
\end{table}
}

\textbf{Evaluation.} To evaluate the class generalisation capabilities of our model, we conduct experiments on both the COCO~\cite{coco} and COCO-NOVEL~\cite{coco-novel} benchmarks. This setup allows us to assess the model's performance on known classes (COCO) and its ability to generalise to unseen classes (COCO-NOVEL).
Note that in the COCO-NOVEL dataset, the model is trained exclusively on the \textit{base} classes and evaluated on \textit{novel} classes, which remain unseen throughout training.

\begin{figure*}[h]
\begin{center}
\includegraphics[width=0.986\textwidth]{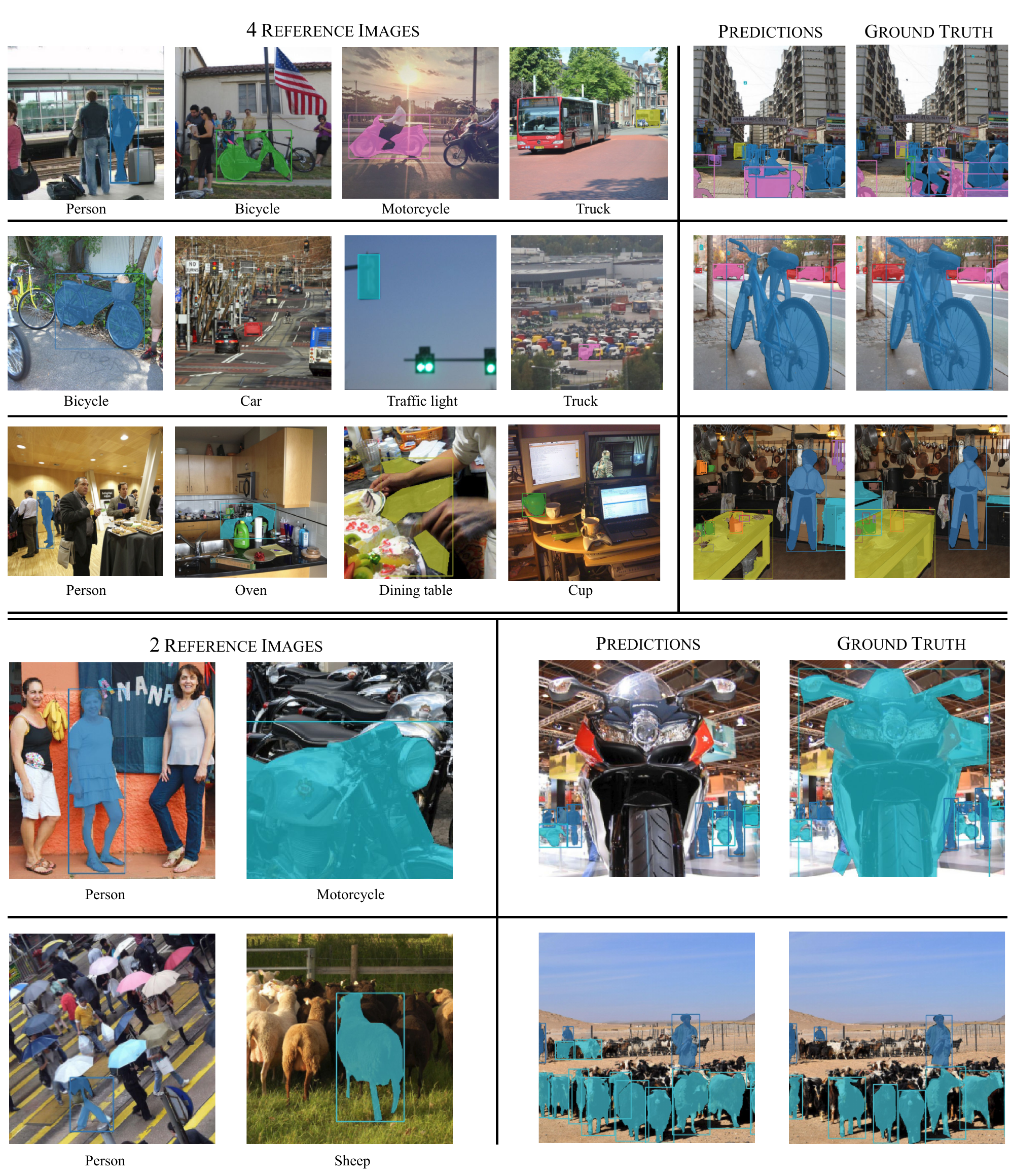}
\end{center}
   \caption{Qualitative visualisations on COCO images. Reference images are used to condition our model for in-context semantic prompting. We run our model per reference image and aggregate results in the visualisation for the target image. Best viewed when zoomed in. By incorporating DETR decoder we natively support multi-instance detection for a given category}
\label{fig:coco-images}
\end{figure*}

\textbf{Results.} Table~\ref{tab:coco} shows that our model achieves competitive segmentation performance on COCO, successfully encoding the semantics of the classes it is trained on. Furthermore, we show qualitative results in Figure \ref{fig:coco-images}, comparing the ground truth annotations and the predicted annotations.

\begin{figure}[h]
    \centering
    \includegraphics[width=\linewidth]{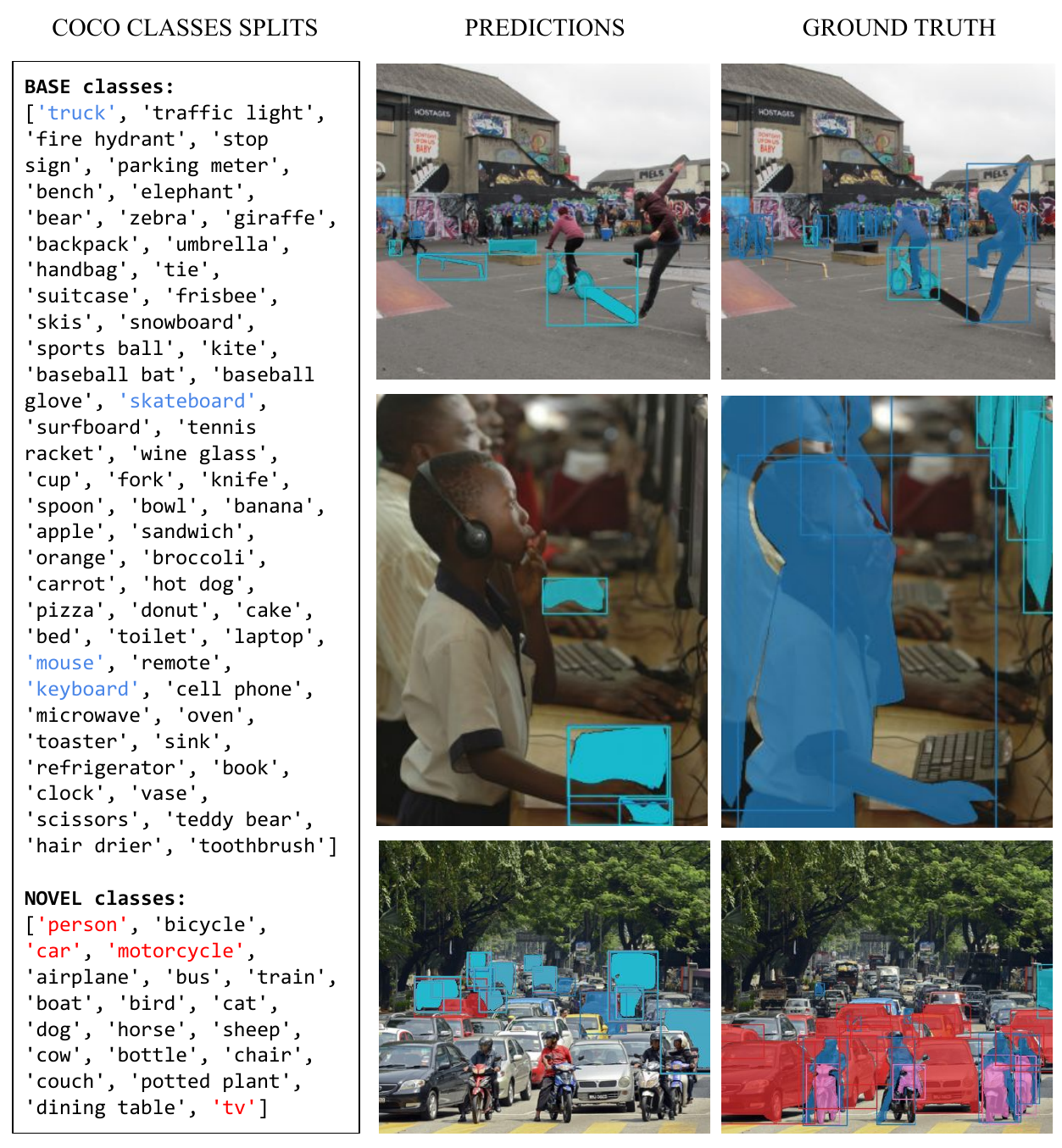}
    \caption{Failure cases for NOVEL classes (unseen during training). The lack of generalisability of our adapted SAM is apparent on unseen semantic categories. We provide the list of categories for both BASE and NOVEL splits for reference. We highlight some base and novel classes to draw the attention of the reader to the corresponding failure cases in the images. In the top example, skateboards from the BASE set are accurately segmented, but persons from the NOVEL set are not. Middle: mouse and keyboard instances from BASE are identified but person and tv from NOVEL are not. Bottom: Trucks from BASE are identified, but car, motorcycle and person from NOVEL are not.}
    \label{fig:coco-novel-images}
\end{figure}

However, when our model is trained only a subset of classes (base classes), it fails to generalise to novel classes. The results obtained are shown in  Table~\ref{tab:coco-novel}, highlighting the large validation mAP gap between base classes and novel classes. Qualitative failure cases are illustrated in Figure~\ref{fig:coco-novel-images}.

\begin{figure}[h]
    \centering
    \includegraphics[width=\linewidth]{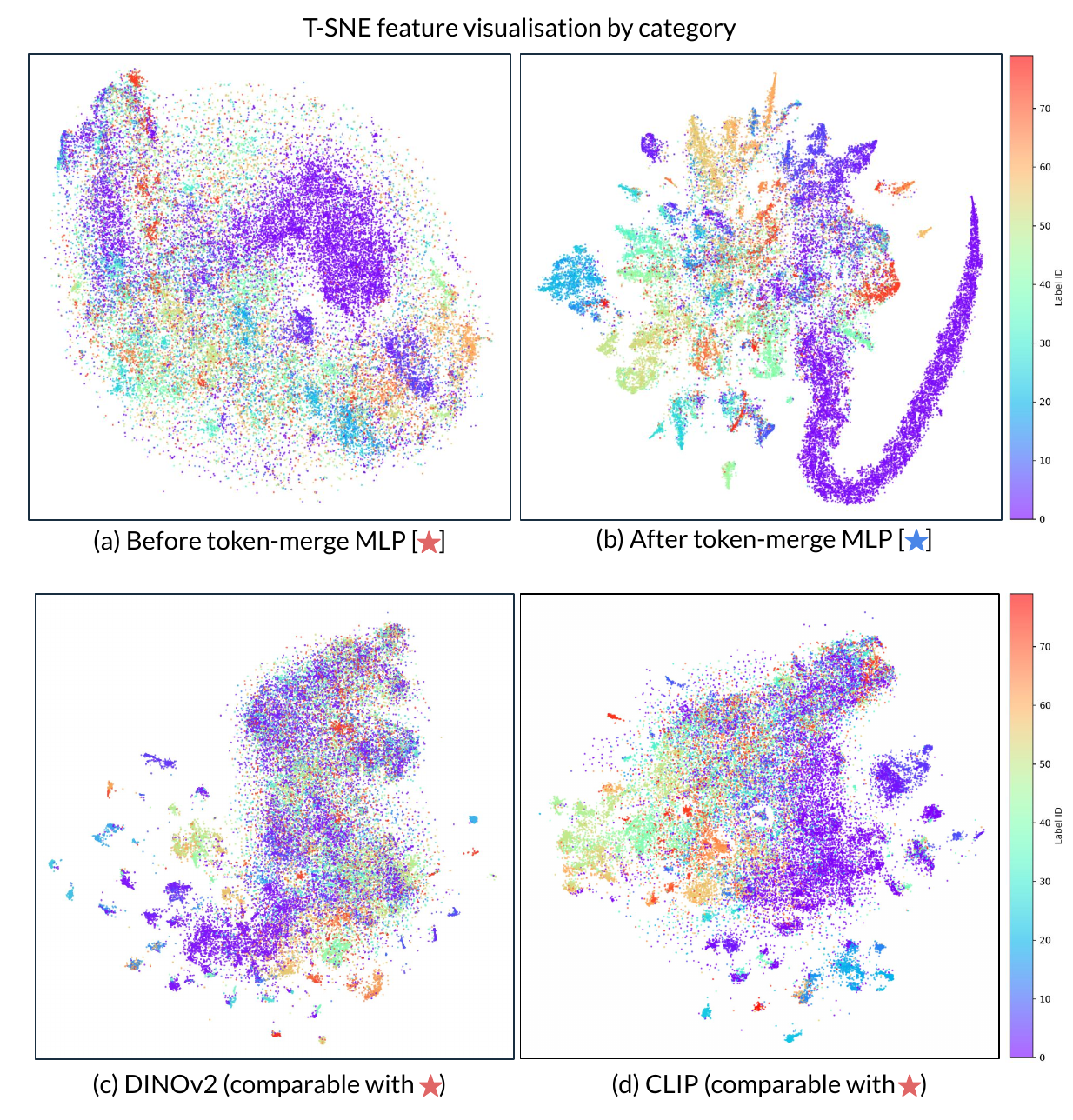}
    \caption{T-SNE visualisation on the COCO trained model features. We refer the reader to Figure \ref{fig:sam-incole} to visualise the locations (red and blue stars) where the feature representations are extracted for T-SNE analysis. (a), (c), (d) correspond to the t-SNE plots on the pretrained frozen representations of SAM~\cite{sam}, DINOv2~\cite{dinov2}, and CLIP~\cite{clip} respectively. It can be visually observed that DINOv2 and CLIP have better class separation (thus, more discriminant features) than SAM. (a), (b) are two different t-SNE visualisations comparing before and after the token-merge MLP, highlighting that semantics are learned only after fine-tuning, thus, limited to the training categories.}
    \label{fig:tsne-feature-visualisation}
\end{figure}

To further investigate this limitation, we analyse the model's internal representations by extracting its latent features before and after the \textit{token-merge MLP} layer (see Figure \ref{fig:sam-incole} marked with red and blue arrows for the location of the features extracted before and after the MLP, respectively). We make use of t-SNE~\cite{tsne} plots to visualise the features in Figure \ref{fig:tsne-feature-visualisation}. These plots reveal that the class semantic information is not present before the MLP. Instead, class semantics are actually being learned in the MLP layer, therefore overfitting to the classes seen during training. While the MLP layer encodes class information, this semantic structure is absent in the earlier features, highlighting that the semantics that are recovered in the model are specific to the classes we tune on, and that these class-specific cues do not generalise to unseen classes.

\textbf{Discussion}. Although our model architecture is able to effectively learn to segment target images based on a semantically meaningful segmentation prompt, we observe the lack of generalisation to unseen classes after the fine-tuning phase. These findings suggest a fundamental limitation: SAM models lack the inherent semantic flexibility required to generalise to novel classes.

\section{Injecting Semantics into SAM}

We have seen that SAM lacks semantics (1) for generic visual understanding tasks (Section \ref{sec:quantify-semantics}), and that (2) even after being finetuned it fails to generalise to novel classes (Section \ref{sec:recovering-semantics}). We now proceed to (3) injecting semantics from external sources.

Motivated by the recent InsDet method~\cite{shen2024high}, we develop a simple, training-free baseline that combines SAM’s segmentation capabilities with the semantic richness of DINOv2 features. 

\textbf{Setup.} As previously defined in Section~\ref{sec:in-context-segmentation-prompting}, in an in-context segmentation prompting setting we have a small subset of reference images for each category. We aim to find instances of such categories in the target images.

\textbf{Evaluation.} The approach consists of three main steps. First, we generate object proposals for all regions in the test image by prompting SAM with a dense grid of points. Then, we use DINOv2 to extract the latent feature representations for all reference images, and the target image, respectively. Lastly, we match the DINOv2 features of the target image with all the encoded reference images features via cosine similarity. We apply simple post-processing steps to remove duplicate predictions for final instance detection.

{\begin{table}[h]
\centering
\setlength{\tabcolsep}{1pt} %
\renewcommand{\arraystretch}{1.2} %

\begin{tabular}{lcccc}
\toprule
\multicolumn{1}{c}{\textbf{COCO}} & \textbf{AP$_{\text{box}}$} & AR$_{\text{box}}$ & \textbf{AP$_{\text{mask}}$} & AR$_{\text{mask}}$ \\
\midrule
SAM+DINOv2 & 13.3 & 39.9 & 11.3 & 34.1 \\
\bottomrule
\end{tabular}
\caption{Preliminary box and mask AP and AR results on COCO validation split using DINOv2 with SAM. Since it is a training-free method, these results can be interpreted as generalisation to unseen classes.}
\label{tab:sam-dinov2}
\end{table}
}

\textbf{Results.} By adopting this approach, our baseline achieves semantic matching at instance level through feature-based similarity rather than extensive training. Table~\ref{tab:sam-dinov2} shows the results obtained on COCO~\cite{coco} validation split. Furthermore, being a fully training-free method makes it naturally adaptable to novel classes (without any risk of overfitting to seen categories). This approach serves as an initial exploration of a semantic injection from external sources. Unlike prior attempts in Section~\ref{sec:related-work}, which required model fine-tuning, this baseline demonstrates that a training-free, feature-based approach can yield promising results. Our preliminary results in this direction indicate a viable signal for improvement, suggesting that incorporating external semantic sources like DINOv2 can unlock SAM’s capacity for better class differentiation, ultimately bridging SAM’s segmentation abilities with meaningful semantic understanding.

\section{Conclusion}

This work has explored SAM as a visual image encoder. Specifically, we examined whether SAM has any class-level information, and if this is strong enough to support tasks that require strong semantic understanding.

Our findings empirically demonstrate that training for label-agnostic mask generation, even with extensive datasets and diverse image distributions, does not inherently endow models with a semantic understanding of distinct object classes. Rather than learning meaningful, object-specific representations, SAM models primarily capture generic features related to ``objectness" and their boundaries---effectively delineating objects without differentiating between specific categories. SAM's core capability lies in spatial region delineation, with any semantic understanding present being weak and difficult to harness effectively.

Furthermore, we have shown that attempts to learn semantics directly on top of SAM lead to overfitting to the training classes, limiting the model’s ability to generalise to unseen classes. These insights underscore the need for external sources (ie semantically-rich backbones) to inject missing semantic information. We thus explore a training-free approach that combines SAM’s segmentation capabilities with the semantic depth of DINOv2 features. We leveraged cosine similarity to improve semantic understanding for in-context instance segmentation. Preliminary results demonstrated that integrating DINOv2 features enabled better distinction between object classes, even for novel categories. This promising direction suggests that incorporating external semantic sources like DINOv2 can significantly enhance SAM’s utility for complex visual tasks, paving the way for future improvements.

\section{Acknowledgements}

Miguel Espinosa was supported through a Centre for Satellite Data in Environmental Science (SENSE) CDT studentship (NE/T00939X/1). This work used JASMIN, the UK's collaborative data analysis environment \url{https://jasmin.ac.uk} \cite{jasmin}.

{
    \small
    \bibliographystyle{ieeenat_fullname}
    \bibliography{main}

\begin{thebibliography}{62}
\providecommand{\natexlab}[1]{#1}
\providecommand{\url}[1]{\texttt{#1}}
\expandafter\ifx\csname urlstyle\endcsname\relax
  \providecommand{\doi}[1]{doi: #1}\else
  \providecommand{\doi}{doi: \begingroup \urlstyle{rm}\Url}\fi

\bibitem[Bai et~al.(2024)Bai, Geng, Mangalam, Bar, Yuille, Darrell, Malik, and Efros]{Bai2023Dec_sequential_modeling}
Yutong Bai, Xinyang Geng, Karttikeya Mangalam, Amir Bar, Alan Yuille, Trevor Darrell, Jitendra Malik, and Alexei~A. Efros.
\newblock {Sequential Modeling Enables Scalable Learning for Large Vision Models}.
\newblock \emph{CVPR}, 2024.

\bibitem[Bar et~al.(2022)Bar, Gandelsman, Darrell, Globerson, and Efros]{Bar2022Sep_visual_icl}
Amir Bar, Yossi Gandelsman, Trevor Darrell, Amir Globerson, and Alexei~A. Efros.
\newblock {Visual Prompting Via Image Inpainting}.
\newblock In \emph{NeurIPS}, 2022.

\bibitem[Brown et~al.(2020)Brown, Mann, Ryder, Subbiah, Kaplan, Dhariwal, Neelakantan, Shyam, Sastry, Askell, et~al.]{brown2020language}
Tom Brown, Benjamin Mann, Nick Ryder, Melanie Subbiah, Jared~D Kaplan, Prafulla Dhariwal, Arvind Neelakantan, Pranav Shyam, Girish Sastry, Amanda Askell, et~al.
\newblock Language models are few-shot learners.
\newblock \emph{NeurIPS}, 2020.

\bibitem[Carion et~al.(2020)Carion, Massa, Synnaeve, Usunier, Kirillov, and Zagoruyko]{detr}
Nicolas Carion, Francisco Massa, Gabriel Synnaeve, Nicolas Usunier, Alexander Kirillov, and Sergey Zagoruyko.
\newblock End-to-end object detection with transformers.
\newblock In \emph{ECCV}, 2020.

\bibitem[Chen et~al.(2019)Chen, Wang, Pang, Cao, Xiong, Li, Sun, Feng, Liu, Xu, Zhang, Cheng, Zhu, Cheng, Zhao, Li, Lu, Zhu, Wu, Dai, Wang, Shi, Ouyang, Loy, and Lin]{mmdetection}
Kai Chen, Jiaqi Wang, Jiangmiao Pang, Yuhang Cao, Yu Xiong, Xiaoxiao Li, Shuyang Sun, Wansen Feng, Ziwei Liu, Jiarui Xu, Zheng Zhang, Dazhi Cheng, Chenchen Zhu, Tianheng Cheng, Qijie Zhao, Buyu Li, Xin Lu, Rui Zhu, Yue Wu, Jifeng Dai, Jingdong Wang, Jianping Shi, Wanli Ouyang, Chen~Change Loy, and Dahua Lin.
\newblock {MMDetection}: Open mmlab detection toolbox and benchmark.
\newblock \emph{arXiv preprint arXiv:1906.07155}, 2019.

\bibitem[Chen et~al.(2023{\natexlab{a}})Chen, Mai, Li, and Chao]{chen2023segment}
Tianle Chen, Zheda Mai, Ruiwen Li, and Wei-lun Chao.
\newblock Segment anything model (sam) enhanced pseudo labels for weakly supervised semantic segmentation.
\newblock \emph{NeurIPS Workshop}, 2023{\natexlab{a}}.

\bibitem[Chen et~al.(2023{\natexlab{b}})Chen, Jia, and Wu]{chen2023fine}
Xu Chen, Yunde Jia, and Yuwei Wu.
\newblock Fine-grained annotation for face anti-spoofing.
\newblock \emph{arXiv preprint arXiv:2310.08142}, 2023{\natexlab{b}}.

\bibitem[Cheng et~al.(2022)Cheng, Misra, Schwing, Kirillov, and Girdhar]{mask2former}
Bowen Cheng, Ishan Misra, Alexander~G. Schwing, Alexander Kirillov, and Rohit Girdhar.
\newblock Masked-attention mask transformer for universal image segmentation.
\newblock In \emph{CVPR}, 2022.

\bibitem[Dosovitskiy et~al.(2021)Dosovitskiy, Beyer, Kolesnikov, Weissenborn, Zhai, Unterthiner, Dehghani, Minderer, Heigold, Gelly, et~al.]{vit}
Alexey Dosovitskiy, Lucas Beyer, Alexander Kolesnikov, Dirk Weissenborn, Xiaohua Zhai, Thomas Unterthiner, Mostafa Dehghani, Matthias Minderer, Georg Heigold, Sylvain Gelly, et~al.
\newblock An image is worth 16x16 words: Transformers for image recognition at scale.
\newblock In \emph{ICLR}, 2021.

\bibitem[Feng et~al.(2021)Feng, Haase-Schütz, Rosenbaum, Hertlein, Gläser, Timm, Wiesbeck, and Dietmayer]{drivingsegment}
Di Feng, Christian Haase-Schütz, Lars Rosenbaum, Heinz Hertlein, Claudius Gläser, Fabian Timm, Werner Wiesbeck, and Klaus Dietmayer.
\newblock Deep multi-modal object detection and semantic segmentation for autonomous driving: Datasets, methods, and challenges.
\newblock \emph{IEEE Transactions on Intelligent Transportation Systems}, 2021.

\bibitem[Gallagher et~al.(2024)Gallagher, Gogia, and Oughton]{gallagher2024multispectral}
James~E Gallagher, Aryav Gogia, and Edward~J Oughton.
\newblock A multispectral automated transfer technique (matt) for machine-driven image labeling utilizing the segment anything model (sam).
\newblock \emph{arXiv preprint arXiv:2402.11413}, 2024.

\bibitem[Girshick(2015)]{FastRCNN}
Ross Girshick.
\newblock Fast {R}-{CNN}.
\newblock In \emph{{ICCV}}, 2015.

\bibitem[Girshick et~al.(2013)Girshick, Donahue, Darrell, and Malik]{RCNN}
Ross~B. Girshick, Jeff Donahue, Trevor Darrell, and Jitendra Malik.
\newblock Rich feature hierarchies for accurate object detection and semantic segmentation.
\newblock In \emph{{CVPR}}, 2013.

\bibitem[He et~al.(2017)He, Gkioxari, Dollár, and Girshick]{MaskRCNN}
Kaiming He, Georgia Gkioxari, Piotr Dollár, and Ross Girshick.
\newblock Mask r-cnn.
\newblock In \emph{ICCV}, 2017.

\bibitem[He et~al.(2022)He, Chen, Xie, Li, Doll{\'a}r, and Girshick]{he2022masked}
Kaiming He, Xinlei Chen, Saining Xie, Yanghao Li, Piotr Doll{\'a}r, and Ross Girshick.
\newblock Masked autoencoders are scalable vision learners.
\newblock In \emph{CVPR}, 2022.

\bibitem[Huang et~al.(2024)Huang, Jiang, Zhang, Qiu, Lu, Lu, and Xing]{Huang2024Jan_learning_to_prompt}
Jiaxing Huang, Kai Jiang, Jingyi Zhang, Han Qiu, Lewei Lu, Shijian Lu, and Eric Xing.
\newblock {Learning to Prompt Segment Anything Models}.
\newblock \emph{ICLR}, 2024.

\bibitem[Jiang and Yang(2023)]{jiang2023segment}
Peng-Tao Jiang and Yuqi Yang.
\newblock Segment anything is a good pseudo-label generator for weakly supervised semantic segmentation.
\newblock \emph{arXiv preprint arXiv:2305.01275}, 2023.

\bibitem[Jiang et~al.(2024)Jiang, Li, Zeng, Ren, Liu, and Zhang]{t_rex2}
Qing Jiang, Feng Li, Zhaoyang Zeng, Tianhe Ren, Shilong Liu, and Lei Zhang.
\newblock {T-Rex2: Towards Generic Object Detection via Text-Visual Prompt Synergy}.
\newblock \emph{ECCV}, 2024.

\bibitem[Kang et~al.(2019)Kang, Liu, Wang, Yu, Feng, and Darrell]{coco-novel}
Bingyi Kang, Zhuang Liu, Xin Wang, Fisher Yu, Jiashi Feng, and Trevor Darrell.
\newblock Few-shot object detection via feature reweighting.
\newblock In \emph{ICCV}, 2019.

\bibitem[Khani et~al.(2024)Khani, Asgari, Sanghi, Amiri, and Hamarneh]{Khani2023Oct_slime}
Aliasghar Khani, Saeid Asgari, Aditya Sanghi, Ali~Mahdavi Amiri, and Ghassan Hamarneh.
\newblock {SLiMe: Segment Like Me}.
\newblock In \emph{ICLR}, 2024.

\bibitem[Kirillov et~al.(2023)Kirillov, Mintun, Ravi, Mao, Rolland, Gustafson, Xiao, Whitehead, Berg, Lo, Doll{\'a}r, and Girshick]{sam}
Alexander Kirillov, Eric Mintun, Nikhila Ravi, Hanzi Mao, Chloe Rolland, Laura Gustafson, Tete Xiao, Spencer Whitehead, Alexander~C. Berg, Wan-Yen Lo, Piotr Doll{\'a}r, and Ross Girshick.
\newblock Segment anything.
\newblock \emph{ICCV}, 2023.

\bibitem[Kulkarni et~al.(2024)Kulkarni, Kanhere, Savani, Chan, Chatterjee, Yi, and Parekh]{kulkarni2024anytime}
Pranav Kulkarni, Adway Kanhere, Dharmam Savani, Andrew Chan, Devina Chatterjee, Paul~H Yi, and Vishwa~S Parekh.
\newblock Anytime, anywhere, anyone: Investigating the feasibility of segment anything model for crowd-sourcing medical image annotations.
\newblock \emph{MIDL Short Papers}, 2024.

\bibitem[Lai et~al.(2024)Lai, Tian, Chen, Li, Yuan, Liu, and Jia]{lai2023lisa}
Xin Lai, Zhuotao Tian, Yukang Chen, Yanwei Li, Yuhui Yuan, Shu Liu, and Jiaya Jia.
\newblock Lisa: Reasoning segmentation via large language model.
\newblock \emph{CVPR}, 2024.

\bibitem[Lawrence et~al.(2013)Lawrence, Bennett, Churchill, Juckes, Kershaw, Pascoe, Pepler, Pritchard, and Stephens]{jasmin}
Bryan~N. Lawrence, Victoria~L. Bennett, James Churchill, Martin Juckes, Philip Kershaw, Stephen Pascoe, Sam Pepler, Matthew Pritchard, and Ag Stephens.
\newblock Storing and manipulating environmental big data with jasmin.
\newblock In \emph{IEEE Big Data}, pages 1--5, San Francisco, 2013. IEEE.

\bibitem[Li et~al.(2024{\natexlab{a}})Li, Jiang, Zhang, Ren, Liu, Zou, Xu, Li, Li, Yang, et~al.]{li2023visual_icl}
Feng Li, Qing Jiang, Hao Zhang, Tianhe Ren, Shilong Liu, Xueyan Zou, Huaizhe Xu, Hongyang Li, Chunyuan Li, Jianwei Yang, et~al.
\newblock Visual in-context prompting.
\newblock \emph{CVPR}, 2024{\natexlab{a}}.

\bibitem[Li et~al.(2024{\natexlab{b}})Li, Zhang, Sun, Zou, Liu, Yang, Li, Zhang, and Gao]{li2023semantic}
Feng Li, Hao Zhang, Peize Sun, Xueyan Zou, Shilong Liu, Jianwei Yang, Chunyuan Li, Lei Zhang, and Jianfeng Gao.
\newblock Semantic-sam: Segment and recognize anything at any granularity.
\newblock \emph{ECCV}, 2024{\natexlab{b}}.

\bibitem[Li et~al.(2022)Li, Mao, Girshick, and He]{li2022exploring}
Yanghao Li, Hanzi Mao, Ross Girshick, and Kaiming He.
\newblock Exploring plain vision transformer backbones for object detection.
\newblock In \emph{ECCV}, 2022.

\bibitem[Lin et~al.(2014)Lin, Maire, Belongie, Hays, Perona, Ramanan, Doll{\'a}r, and Zitnick]{coco}
Tsung-Yi Lin, Michael Maire, Serge Belongie, James Hays, Pietro Perona, Deva Ramanan, Piotr Doll{\'a}r, and C~Lawrence Zitnick.
\newblock Microsoft coco: Common objects in context.
\newblock In \emph{ECCV}, 2014.

\bibitem[Lin et~al.(2017)Lin, Goyal, Girshick, He, and Doll{\'a}r]{focal_loss}
Tsung-Yi Lin, Priya Goyal, Ross Girshick, Kaiming He, and Piotr Doll{\'a}r.
\newblock Focal loss for dense object detection.
\newblock In \emph{ICCV}, 2017.

\bibitem[Liu et~al.(2024{\natexlab{a}})Liu, Jing, Li, Zhu, Chen, Wang, and Shen]{liu2024simple}
Yang Liu, Chenchen Jing, Hengtao Li, Muzhi Zhu, Hao Chen, Xinlong Wang, and Chunhua Shen.
\newblock A simple image segmentation framework via in-context examples.
\newblock \emph{NeurIPS}, 2024{\natexlab{a}}.

\bibitem[Liu et~al.(2024{\natexlab{b}})Liu, Zhu, Li, Chen, Wang, and Shen]{Liu2023May_matcher}
Yang Liu, Muzhi Zhu, Hengtao Li, Hao Chen, Xinlong Wang, and Chunhua Shen.
\newblock {Matcher: Segment Anything with One Shot Using All-Purpose Feature Matching}.
\newblock In \emph{ICLR}, 2024{\natexlab{b}}.

\bibitem[Long et~al.(2015)Long, Shelhamer, and Darrell]{fcn}
Jonathan Long, Evan Shelhamer, and Trevor Darrell.
\newblock Fully convolutional networks for semantic segmentation.
\newblock In \emph{CVPR}, 2015.

\bibitem[Milletari et~al.(2016)Milletari, Navab, and Ahmadi]{dice_loss}
Fausto Milletari, Nassir Navab, and Seyed-Ahmad Ahmadi.
\newblock V-net: Fully convolutional neural networks for volumetric medical image segmentation.
\newblock In \emph{3DV}, 2016.

\bibitem[Nguyen et~al.(2023)Nguyen, Li, Ojha, and Lee]{Nguyen2023Nov_visual_instr_inversion}
Thao Nguyen, Yuheng Li, Utkarsh Ojha, and Yong~Jae Lee.
\newblock {Visual Instruction Inversion: Image Editing via Image Prompting}.
\newblock In \emph{NeurIPS}, 2023.

\bibitem[Oquab et~al.(2023)Oquab, Darcet, Moutakanni, Vo, Szafraniec, Khalidov, Fernandez, Haziza, Massa, El-Nouby, Howes, Huang, Xu, Sharma, Li, Galuba, Rabbat, Assran, Ballas, Synnaeve, Misra, Jegou, Mairal, Labatut, Joulin, and Bojanowski]{dinov2}
Maxime Oquab, Timothée Darcet, Theo Moutakanni, Huy~V. Vo, Marc Szafraniec, Vasil Khalidov, Pierre Fernandez, Daniel Haziza, Francisco Massa, Alaaeldin El-Nouby, Russell Howes, Po-Yao Huang, Hu Xu, Vasu Sharma, Shang-Wen Li, Wojciech Galuba, Mike Rabbat, Mido Assran, Nicolas Ballas, Gabriel Synnaeve, Ishan Misra, Herve Jegou, Julien Mairal, Patrick Labatut, Armand Joulin, and Piotr Bojanowski.
\newblock Dinov2: Learning robust visual features without supervision.
\newblock \emph{TMLR}, 2023.

\bibitem[Pan et~al.(2024)Pan, Tang, Wang, and Shan]{pan2023tap}
Ting Pan, Lulu Tang, Xinlong Wang, and Shiguang Shan.
\newblock Tokenize anything via prompting.
\newblock \emph{ECCV}, 2024.

\bibitem[Radford et~al.(2021)Radford, Kim, Hallacy, Ramesh, Goh, Agarwal, Sastry, Askell, Mishkin, Clark, et~al.]{clip}
Alec Radford, Jong~Wook Kim, Chris Hallacy, Aditya Ramesh, Gabriel Goh, Sandhini Agarwal, Girish Sastry, Amanda Askell, Pamela Mishkin, Jack Clark, et~al.
\newblock Learning transferable visual models from natural language supervision.
\newblock In \emph{ICML}, 2021.

\bibitem[Ravi et~al.(2024)Ravi, Gabeur, Hu, Hu, Ryali, Ma, Khedr, R{\"a}dle, Rolland, Gustafson, Mintun, Pan, Alwala, Carion, Wu, Girshick, Doll{\'a}r, and Feichtenhofer]{sam2}
Nikhila Ravi, Valentin Gabeur, Yuan-Ting Hu, Ronghang Hu, Chaitanya Ryali, Tengyu Ma, Haitham Khedr, Roman R{\"a}dle, Chloe Rolland, Laura Gustafson, Eric Mintun, Junting Pan, Kalyan~Vasudev Alwala, Nicolas Carion, Chao-Yuan Wu, Ross Girshick, Piotr Doll{\'a}r, and Christoph Feichtenhofer.
\newblock Sam 2: Segment anything in images and videos.
\newblock \emph{arXiv preprint arXiv:2408.00714}, 2024.

\bibitem[Ren et~al.(2015)Ren, He, Girshick, and Sun]{FasterRCNN}
Shaoqing Ren, Kaiming He, Ross Girshick, and Jian Sun.
\newblock Faster {R-CNN}: Towards real-time object detection with region proposal networks.
\newblock In \emph{NeurIPS}, 2015.

\bibitem[Ronneberger et~al.(2015)Ronneberger, Fischer, and Brox]{unet}
Olaf Ronneberger, Philipp Fischer, and Thomas Brox.
\newblock U-net: Convolutional networks for biomedical image segmentation.
\newblock In \emph{MICCAI}, 2015.

\bibitem[Ru et~al.(2023)Ru, Zheng, Zhan, and Du]{ru2023token}
Lixiang Ru, Heliang Zheng, Yibing Zhan, and Bo Du.
\newblock Token contrast for weakly-supervised semantic segmentation.
\newblock In \emph{CVPR}, 2023.

\bibitem[Ryali et~al.(2023)Ryali, Hu, Bolya, Wei, Fan, Huang, Aggarwal, Chowdhury, Poursaeed, Hoffman, Malik, Li, and Feichtenhofer]{ryali2023hiera}
Chaitanya Ryali, Yuan-Ting Hu, Daniel Bolya, Chen Wei, Haoqi Fan, Po-Yao Huang, Vaibhav Aggarwal, Arkabandhu Chowdhury, Omid Poursaeed, Judy Hoffman, Jitendra Malik, Yanghao Li, and Christoph Feichtenhofer.
\newblock Hiera: A hierarchical vision transformer without the bells-and-whistles.
\newblock \emph{ICML}, 2023.

\bibitem[Shen et~al.(2024)Shen, Zhao, Kwon, Kim, Li, and Kong]{shen2024high}
Qianqian Shen, Yunhan Zhao, Nahyun Kwon, Jeeeun Kim, Yanan Li, and Shu Kong.
\newblock A high-resolution dataset for instance detection with multi-view object capture.
\newblock \emph{NeurIPS}, 2024.

\bibitem[Strudel et~al.(2021)Strudel, Garcia, Laptev, and Schmid]{segmenter}
Robin Strudel, Ricardo Garcia, Ivan Laptev, and Cordelia Schmid.
\newblock Segmenter: Transformer for semantic segmentation.
\newblock In \emph{ICCV}, 2021.

\bibitem[Sun et~al.(2023)Sun, Chen, Wang, Wang, and Li]{Sun2023Apr_exploring_effective_factors}
Yanpeng Sun, Qiang Chen, Jian Wang, Jingdong Wang, and Zechao Li.
\newblock {Exploring Effective Factors for Improving Visual In-Context Learning}.
\newblock \emph{arXiv}, 2023.

\bibitem[Sun et~al.(2024)Sun, Chen, Zhang, Zhang, Chen, Zhang, Ding, Wang, and Li]{sun2024vrp}
Yanpeng Sun, Jiahui Chen, Shan Zhang, Xinyu Zhang, Qiang Chen, Gang Zhang, Errui Ding, Jingdong Wang, and Zechao Li.
\newblock Vrp-sam: Sam with visual reference prompt.
\newblock \emph{CVPR}, 2024.

\bibitem[Tancik et~al.(2020)Tancik, Srinivasan, Mildenhall, Fridovich-Keil, Raghavan, Singhal, Ramamoorthi, Barron, and Ng]{tancik2020fourier}
Matthew Tancik, Pratul Srinivasan, Ben Mildenhall, Sara Fridovich-Keil, Nithin Raghavan, Utkarsh Singhal, Ravi Ramamoorthi, Jonathan Barron, and Ren Ng.
\newblock Fourier features let networks learn high frequency functions in low dimensional domains.
\newblock \emph{NeurIPS}, 2020.

\bibitem[Touvron et~al.(2021)Touvron, Cord, Douze, Massa, Sablayrolles, and J{\'e}gou]{deit}
Hugo Touvron, Matthieu Cord, Matthijs Douze, Francisco Massa, Alexandre Sablayrolles, and Herv{\'e} J{\'e}gou.
\newblock Training data-efficient image transformers \& distillation through attention.
\newblock In \emph{ICML}, 2021.

\bibitem[van~der Maaten and Hinton(2008)]{tsne}
Laurens van~der Maaten and Geoffrey Hinton.
\newblock Visualizing data using t-sne.
\newblock \emph{JMLR}, 2008.

\bibitem[Wang et~al.(2023{\natexlab{a}})Wang, Wang, Cao, Shen, and Huang]{Painter}
Xinlong Wang, Wen Wang, Yue Cao, Chunhua Shen, and Tiejun Huang.
\newblock Images speak in images: A generalist painter for in-context visual learning.
\newblock In \emph{CVPR}, 2023{\natexlab{a}}.

\bibitem[Wang et~al.(2023{\natexlab{b}})Wang, Zhang, Cao, Wang, Shen, and Huang]{SegGPT}
Xinlong Wang, Xiaosong Zhang, Yue Cao, Wen Wang, Chunhua Shen, and Tiejun Huang.
\newblock Seggpt: Segmenting everything in context.
\newblock In \emph{ICCV}, 2023{\natexlab{b}}.

\bibitem[Xie et~al.(2023)Xie, Wang, Ma, Chen, Lu, Yang, Shi, and Lin]{xie2023edit}
Defeng Xie, Ruichen Wang, Jian Ma, Chen Chen, Haonan Lu, Dong Yang, Fobo Shi, and Xiaodong Lin.
\newblock Edit everything: A text-guided generative system for images editing.
\newblock \emph{ACMMM}, 2023.

\bibitem[Yang et~al.(2023)Yang, Gao, Li, Gao, Wang, and Zheng]{yang2023track}
Jinyu Yang, Mingqi Gao, Zhe Li, Shang Gao, Fangjing Wang, and Feng Zheng.
\newblock Track anything: Segment anything meets videos.
\newblock \emph{arXiv preprint arXiv:2304.11968}, 2023.

\bibitem[Yao et~al.(2024)Yao, Wang, Ye, and Liu]{yao2024matte}
Jingfeng Yao, Xinggang Wang, Lang Ye, and Wenyu Liu.
\newblock Matte anything: Interactive natural image matting with segment anything model.
\newblock \emph{Image and Vision Computing}, 2024.

\bibitem[Yu et~al.(2023)Yu, Feng, Feng, Liu, Jin, Zeng, and Chen]{yu2023inpaint}
Tao Yu, Runseng Feng, Ruoyu Feng, Jinming Liu, Xin Jin, Wenjun Zeng, and Zhibo Chen.
\newblock Inpaint anything: Segment anything meets image inpainting.
\newblock \emph{arXiv preprint arXiv:2304.06790}, 2023.

\bibitem[Zhan et~al.(2023)Zhan, Zheng, Xie, and Zisserman]{zhan2023does}
Guanqi Zhan, Chuanxia Zheng, Weidi Xie, and Andrew Zisserman.
\newblock What does stable diffusion know about the 3d scene?
\newblock \emph{arXiv preprint arXiv:2310.06836}, 2023.

\bibitem[Zhang et~al.(2024{\natexlab{a}})Zhang, Herrmann, Hur, Polania~Cabrera, Jampani, Sun, and Yang]{zhang2024tale}
Junyi Zhang, Charles Herrmann, Junhwa Hur, Luisa Polania~Cabrera, Varun Jampani, Deqing Sun, and Ming-Hsuan Yang.
\newblock A tale of two features: Stable diffusion complements dino for zero-shot semantic correspondence.
\newblock \emph{NeurIPS}, 2024{\natexlab{a}}.

\bibitem[Zhang et~al.(2024{\natexlab{b}})Zhang, Jiang, Guo, Yan, Pan, Ma, Dong, Gao, and Li]{Zhang2023May_personalise_SAM}
Renrui Zhang, Zhengkai Jiang, Ziyu Guo, Shilin Yan, Junting Pan, Xianzheng Ma, Hao Dong, Peng Gao, and Hongsheng Li.
\newblock {Personalize Segment Anything Model with One Shot}.
\newblock In \emph{ICLR}, 2024{\natexlab{b}}.

\bibitem[Zhao et~al.(2021)Zhao, Wallace, Feng, Klein, and Singh]{zhao2021calibrate}
Zihao Zhao, Eric Wallace, Shi Feng, Dan Klein, and Sameer Singh.
\newblock Calibrate before use: Improving few-shot performance of language models.
\newblock In \emph{ICML}, 2021.

\bibitem[Zhu et~al.(2024)Zhu, Liu, Luo, Jing, Chen, Xu, Wang, and Shen]{zhu2024unleashing}
Muzhi Zhu, Yang Liu, Zekai Luo, Chenchen Jing, Hao Chen, Guangkai Xu, Xinlong Wang, and Chunhua Shen.
\newblock Unleashing the potential of the diffusion model in few-shot semantic segmentation.
\newblock \emph{NeurIPS}, 2024.

\bibitem[Zou* et~al.(2023{\natexlab{a}})Zou*, Dou*, Yang*, Gan, Li, Li, Dai, Behl, Wang, Yuan, Peng, Wang, Lee*, and Gao*]{zou2023xdecoder}
Xueyan Zou*, Zi-Yi Dou*, Jianwei Yang*, Zhe Gan, Linjie Li, Chunyuan Li, Xiyang Dai, Harkirat Behl, Jianfeng Wang, Lu Yuan, Nanyun Peng, Lijuan Wang, Yong~Jae Lee*, and Jianfeng Gao*.
\newblock Generalized decoding for pixel, image and language.
\newblock \emph{CVPR}, 2023{\natexlab{a}}.

\bibitem[Zou* et~al.(2023{\natexlab{b}})Zou*, Yang*, Zhang*, Li*, Li, Wang, Wang, Gao*, and Lee*]{zou2023seem}
Xueyan Zou*, Jianwei Yang*, Hao Zhang*, Feng Li*, Linjie Li, Jianfeng Wang, Lijuan Wang, Jianfeng Gao*, and Yong~Jae Lee*.
\newblock Segment everything everywhere all at once.
\newblock \emph{NeurIPS}, 2023{\natexlab{b}}.

\end{thebibliography}
}

\clearpage
\setcounter{page}{1}
\maketitlesupplementary

\section{Linear Probing Training Details}
\label{sec:supp-training}

\begin{figure}[h]
    \centering
    \includegraphics[width=\linewidth]{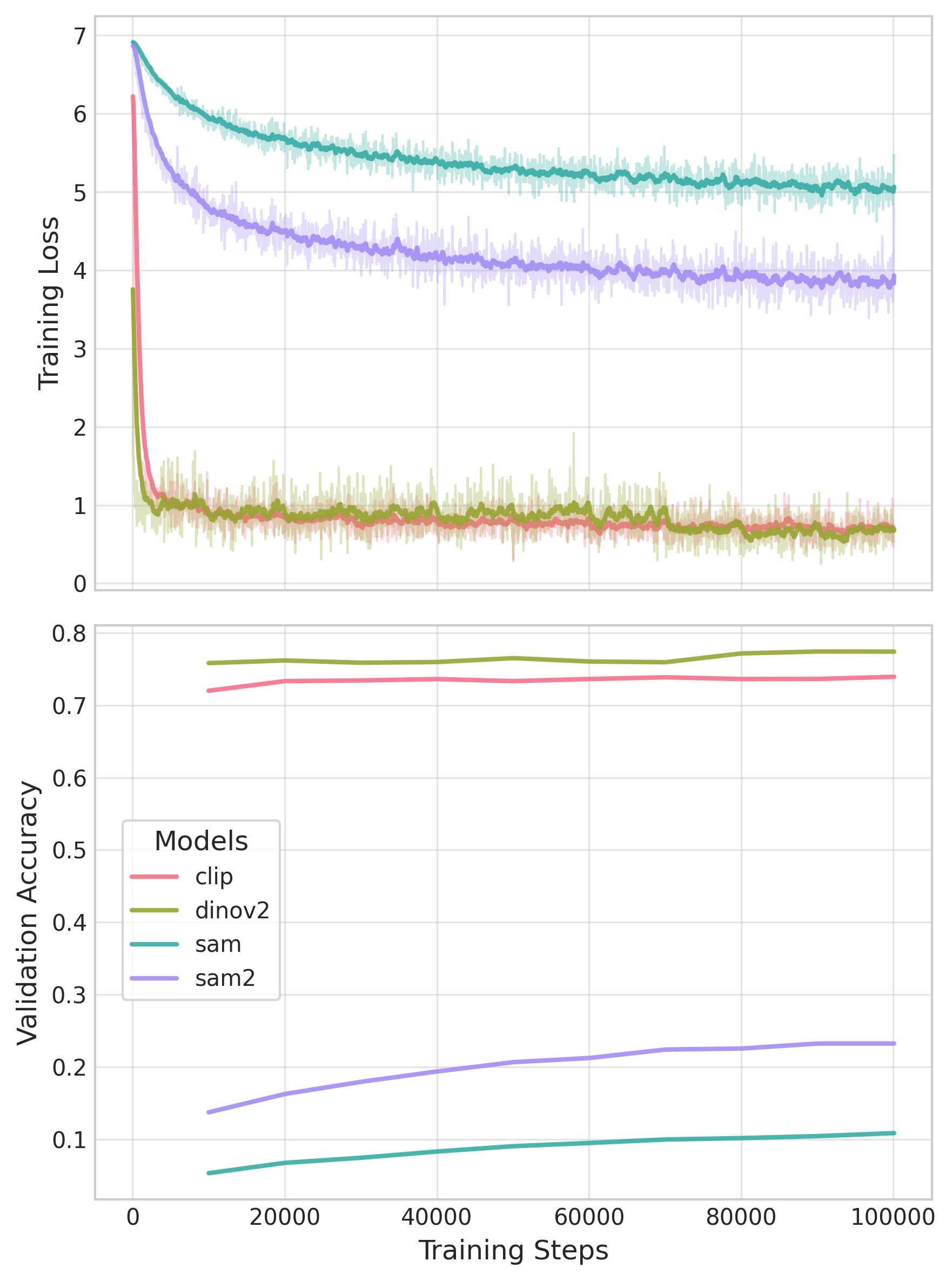}
    \caption{Training loss and validation accuracy curves for SAM, SAM 2, CLIP, and DINOv2 on ImageNet1K image classification using linear probing.}
    \label{fig:supp-training-curves}
\end{figure}

This section provides additional insights into the training process for our experiments. Figure~\ref{fig:supp-training-curves} shows the training loss and validation accuracy curves for SAM, SAM 2, CLIP, and DINOv2 on ImageNet1K classification using linear probing. These plots highlight a significant performance gap between SAM models and established visual encoders such as CLIP and DINOv2, underscoring SAM’s limited semantic discriminability for image classification tasks.

We used the following hyperparameters across all linear probing experiments: learning rate of $10^{-3}$, batch size of 128, 10 epochs, AdamW optimiser, fp16 mixed-precision, input image resolutions were standardised to 1024 for SAM models and 224 for CLIP and DINOv2, following their original configurations.

\section{Codebase and Implementation Details} \label{sec:supp-codebase}

Our implementation builds on the robust and widely used MMDetection framework~\cite{mmdetection}, leveraging its modular design for better experimentation with various architectures and datasets.

Linear Probing: For linear probing experiments, we developed a lightweight Python script to attach and train classification heads on frozen encoders. This standalone script is included in our code repository.

In-context Segmentation Prompting: The architecture for in-context instance segmentation reuses the SAM image encoder and decoder with minimal modifications. We added a DETR decoder and lightweight MLP layers to bridge semantic reference signals with segmentation predictions. Training scripts for this pipeline are also provided.

Training-Free Baseline: The training-free method that combines SAM and DINOv2 involves a grid-point-based proposal generation in SAM, followed by cosine similarity matching using DINOv2 features.

\section{Limitations and Future Directions} \label{sec:supp-limitations}

While this study provides valuable insights, certain limitations warrant further exploration:

Fine-tuning Trade-offs: Although we adapted SAM with lightweight fine-tuning, the lack of generalisation to unseen classes suggests fundamental constraints in the pretrained model.

Semantic Integration: Our training-free approach integrating SAM and DINOv2 shows promise but could benefit from improved alignment mechanisms for reference and target embeddings.

Dataset Diversity: While we focused on COCO and COCO-NOVEL, future work could extend these methods to more diverse datasets to validate broader applicability.

\end{document}